\pdfoutput=1

\documentclass[11pt]{article}

\usepackage[]{acl}

\usepackage{times}
\usepackage{latexsym}
\usepackage{amsmath}
\usepackage[T1]{fontenc}

\usepackage[utf8]{inputenc}

\usepackage{microtype}

\usepackage{inconsolata}

\usepackage{xcolor}
\usepackage{booktabs}
\usepackage{multirow}
\usepackage{graphicx}
\usepackage{graphics}
\usepackage{subfig}
\usepackage{arydshln}
\usepackage{enumitem}
\usepackage{amsmath}
\usepackage{bm}
\usepackage{caption}

\title{A Learning Rate Path Switching Training Paradigm for\\Version Updates of Large Language Models}

\author{Zhihao Wang\textsuperscript{1*}\quad
        Shiyu Liu\textsuperscript{3*}\quad
        Jianheng Huang\textsuperscript{1}\quad
        Zheng Wang\textsuperscript{2}\\
        \textbf{Yixuan Liao\textsuperscript{2}\quad
        Xiaoxin Chen\textsuperscript{2}\quad
        Junfeng Yao\textsuperscript{1}\quad
        Jinsong Su\textsuperscript{1,3,4$\dagger$}}\\
        \textsuperscript{1}School of Informatics, Xiamen University, China\quad
        \textsuperscript{2}vivo AI Lab, China\\
        \textsuperscript{3}Institute of Artificial Intelligence, Xiamen University, China\\     
        \textsuperscript{4}Shanghai Artificial Intelligence Laboratory, China\\
        \texttt{\{zhwang,liushiyu213\}@stu.xmu.edu.cn}\quad\texttt{jssu@xmu.edu.cn}
}

\begin{document}
\maketitle

\renewcommand{\thefootnote}{\fnsymbol{footnote}}
\footnotetext{Work was done when Zhihao Wang, Shiyu Liu and Jianheng Huang were interning at vivo AI Lab.}
\footnotetext[1]{~Equal contribution.}
\footnotetext[2]{~Corresponding author.}
\renewcommand{\thefootnote}{\arabic{footnote}}

\begin{abstract}
Due to the continuous emergence of new data, version updates have become an indispensable requirement for Large Language Models (LLMs).
The training paradigms for version updates of LLMs include pre-training from scratch (PTFS) and continual pre-training (CPT).
Preliminary experiments demonstrate that PTFS achieves better pre-training performance, while CPT has lower training cost.
Moreover, their performance and training cost gaps widen progressively with version updates.
To investigate the underlying reasons for this phenomenon, we analyze the effect of learning rate adjustments during the two stages of CPT: preparing an initialization checkpoint and continual pre-training based on this checkpoint.
We find that a large learning rate in the first stage and a complete learning rate decay process in the second stage are crucial for version updates of LLMs.
Hence, we propose a learning rate path switching training paradigm.
Our paradigm comprises one main path, where we pre-train a LLM with the maximal learning rate, and multiple branching paths, each of which corresponds to an update of the LLM with newly-added training data.
Extensive experiments demonstrate the effectiveness and generalization of our paradigm.
Particularly, when training four versions of LLMs, our paradigm reduces the total training cost to 58\% compared to PTFS, while maintaining comparable pre-training performance.

\end{abstract}

\section{Introduction}

In recent years, there has been significant progress in the research of Large Language Models (LLMs).
By performing large-scale training on massive datasets, LLMs have demonstrated remarkable capabilities, contributing to various fields~\cite{vllm-survey, driving-llm-survey, agent-survey1, agent-survey2}.
However, the training cost of LLMs is significantly higher than that of traditional NLP models.
Particularly, in practical applications, LLMs have to face the need for {\em version updates} due to the continuous emergence of new data, which exacerbates the training cost of LLMs.
Therefore, reducing training cost while maintaining optimal pre-training performance across different versions has become one of the pivotal challenges for LLMs.

Generally, training paradigms applicable for version updates of LLMs can be categorized into two types:
1) {Pre-Training From Scratch (PTFS)}:
retraining new versions of LLMs on both old and new data.
The well-known LLMs including LLaMA~\cite{llama, llama2}, GLM~\cite{glm-130b}, and Baichuan~\cite{baichuan2} are updated via this paradigm.
2) {Continual Pre-Training (CPT)}: 
further pre-training new versions of LLMs on only new data based on the checkpoints from old versions.
This paradigm is often utilized in resource constrained scenarios, such as limited computational resources or unavailability of old data.

In this paper, we firstly conduct preliminary experiments to compare the above two paradigms in version updates of LLMs. 
Compared with PTFS, CPT uses previous checkpoints for initialization, resulting in lower total training cost.
However, CPT suffers from inferior pre-training performance, which becomes increasingly serious as version updates progress.
To study the reasons for this phenomenon, we break down the CPT process into two stages: the first stage involves preparing an initialization checkpoint, and the second stage performing continual pre-training based on this checkpoint.
Then, we conduct two groups of experiments to analyze the effect of learning rate adjustments during these two stages, leading to two conclusions:
1) the larger the learning rate in the first stage, the better the performance of updated LLMs in the second stage;
2) for the second stage, a complete learning rate decay process is beneficial to ensure the optimal performance of updated LLMs.

\begin{figure}[t]
    \begin{center}
        \includegraphics[height=0.28\textheight]{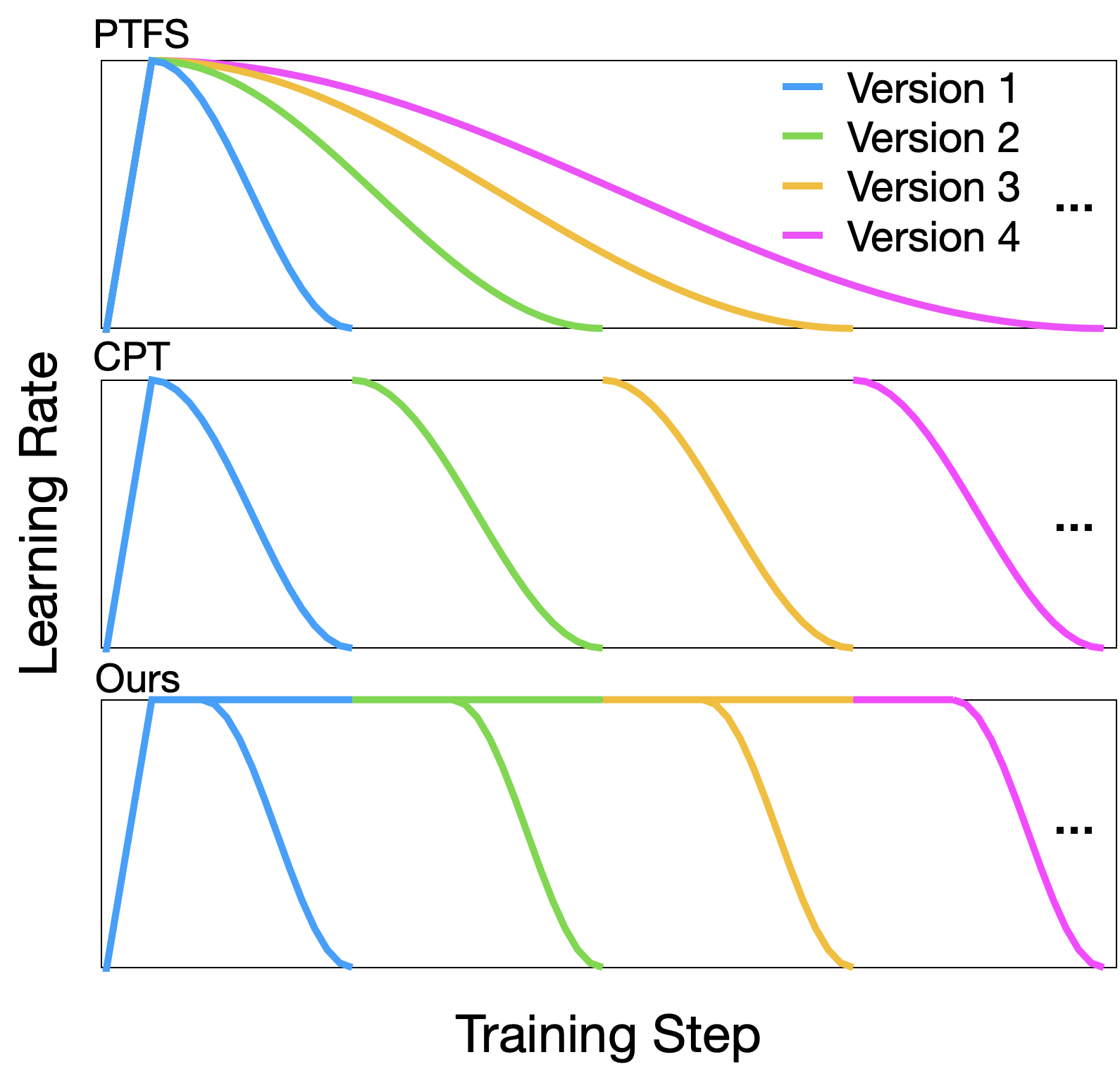}
        \caption{The learning rate curves of cosine learning rate schedule under PTFS, CPT\protect\footnotemark~and our paradigm, all of which are used to update four versions of LLMs. Here, different color curves represent different version updates of LLMs.}
        \label{fig:diff-paradigm}
    \end{center}
\end{figure}

\begin{figure}[t]
    \begin{center}
        \includegraphics[width=0.47\textwidth]{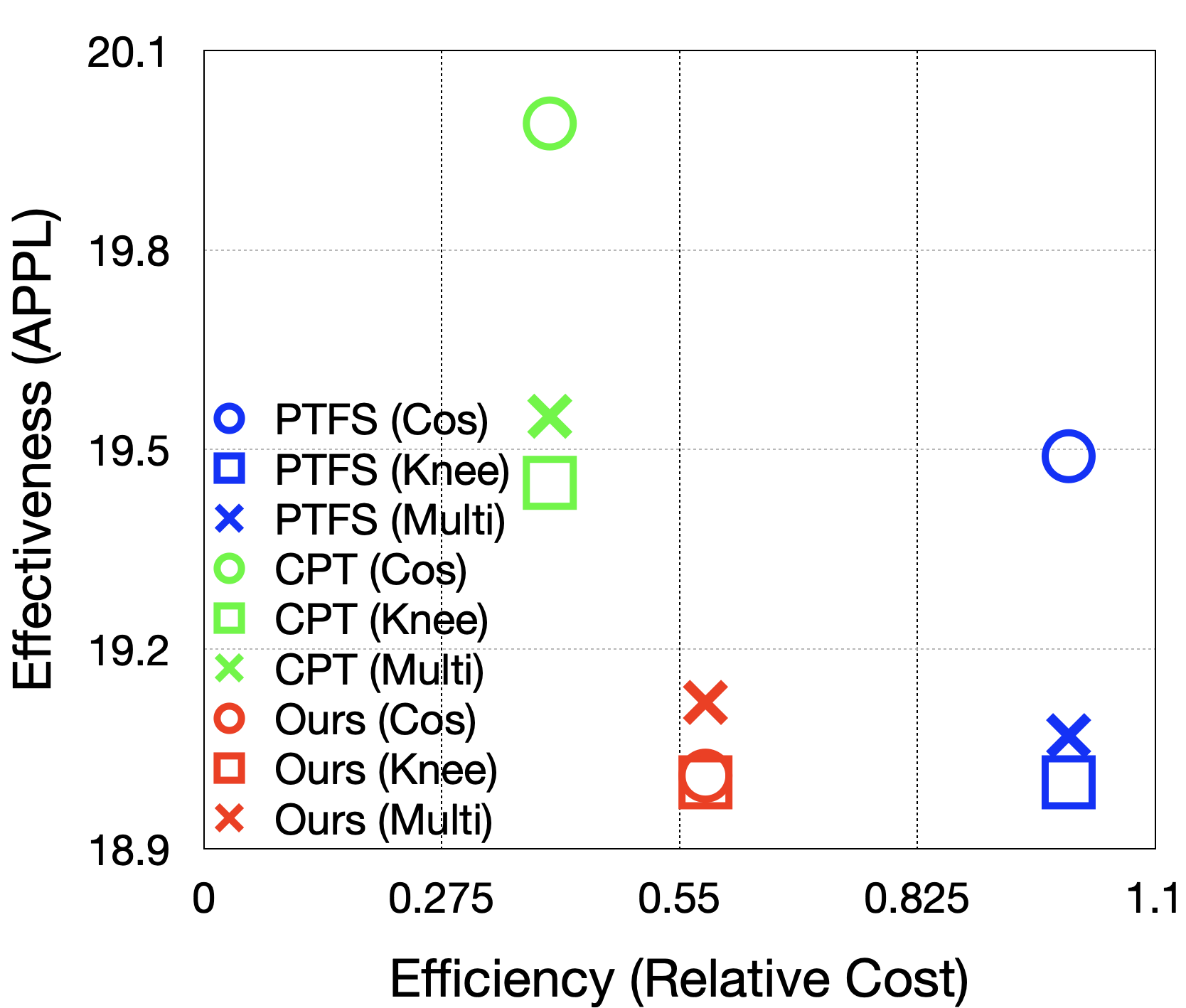}
        \caption{The comparison of different training paradigms. ``APPL'' ($\downarrow$) denotes the average perplexity of LLMs across different versions, ``Relative Cost'' ($\downarrow$) is the ratio of the total training steps across different versions of each paradigm to the total training steps of PTFS. The lower left corner achieves the best trade-off.}
        \label{fig:appl_cost}
    \end{center}
\end{figure}

Based on the above analyses, we propose a learning rate path switching training paradigm for version updates of LLMs.
To better illustrate our paradigm, we take the most commonly used cosine learning rate schedule~\cite{cos} as an example, and plot the learning rate curves of PTFS, CPT and our paradigm in Figure~\ref{fig:diff-paradigm}.
Please note that our paradigm is also applicable to other schedules, such as Knee~\cite{Knee}, and multi-step~\cite{deepseek} learning rate schedules.

\footnotetext{In fact, multiple CPT variants can be used for version updates of LLMs. We compare these variants in Appendix~\ref{sec:opt-lr-set} and retain only the best-performing variant in the subsequent experiments.}

In short, the learning rate curve of our paradigm comprises one main path and multiple branching paths, each of which corresponds to a version update of LLM.
As shown by the main path in Figure~\ref{fig:diff-paradigm}, we pre-train a LLM with the maximal learning rate, providing superior initialization checkpoints for subsequent continual pre-training.
When we want to update the LLM with newly-added training data, we perform continual pre-training on the LLM with a dynamically-adjusted learning rate. 
Referring back to Figure~\ref{fig:diff-paradigm}, after a few steps of training with the maximal learning rate, the learning rate fast decays to its minimum, which effectively ensures the training performance of the updated LLM.
Meanwhile, on the main path, we continue to pre-train the original checkpoint with the maximal learning rate, facilitating subsequent LLM updates.

Our paradigm better balances model performance and training cost compared to the other two paradigms, as detailed in Figure~\ref{fig:appl_cost}.
To summarize, our main contributions are as follows:
\begin{itemize}[leftmargin=*,topsep=0.12em,itemsep=0.12em,parsep=0.12em]
    \item 
    We conduct preliminary experiments to compare PTFS and CPT for version updates of LLMs.
    Furthermore, our in-depth analyses show that initially using a large learning rate and subsequent learning rate decay are crucial for improving the performance of updated LLMs.
    
    \item
    We propose a learning rate path switching paradigm for version updates of LLMs. 
    To the best of our knowledge, our work is the first attempt to explore how to balance model performance and training cost for version updates of LLMs.

    \item Experimental results and in-depth analyses strongly demonstrate the effectiveness and generalization of our paradigm.
    Particularly, when training four versions of LLMs, our paradigm achieves comparable pre-training performance to PTFS with only 58\% of the total training cost.
\end{itemize}

\section{Preliminary Study}
\label{sec:preliminary}

In this section, we first compare the performance of PTFS and CPT in version updates of LLMs, and then analyze the underlying reasons for their performance gap.
\subsection{Setup}

\paragraph{Model} 
In this study, we use LLaMA-1.2B ~\cite{llama,llama2} as our base LLM and train for four versions.
When employing PTFS, the total training steps for these four versions are 10K, 20K, 30K, and 40K, respectively.
For CPT, each LLM update only requires 10K training steps.
We train all LLMs with a batch size of 1.05M tokens.

\begin{figure}[t]
    \begin{center}
        \includegraphics[height=0.26\textheight]{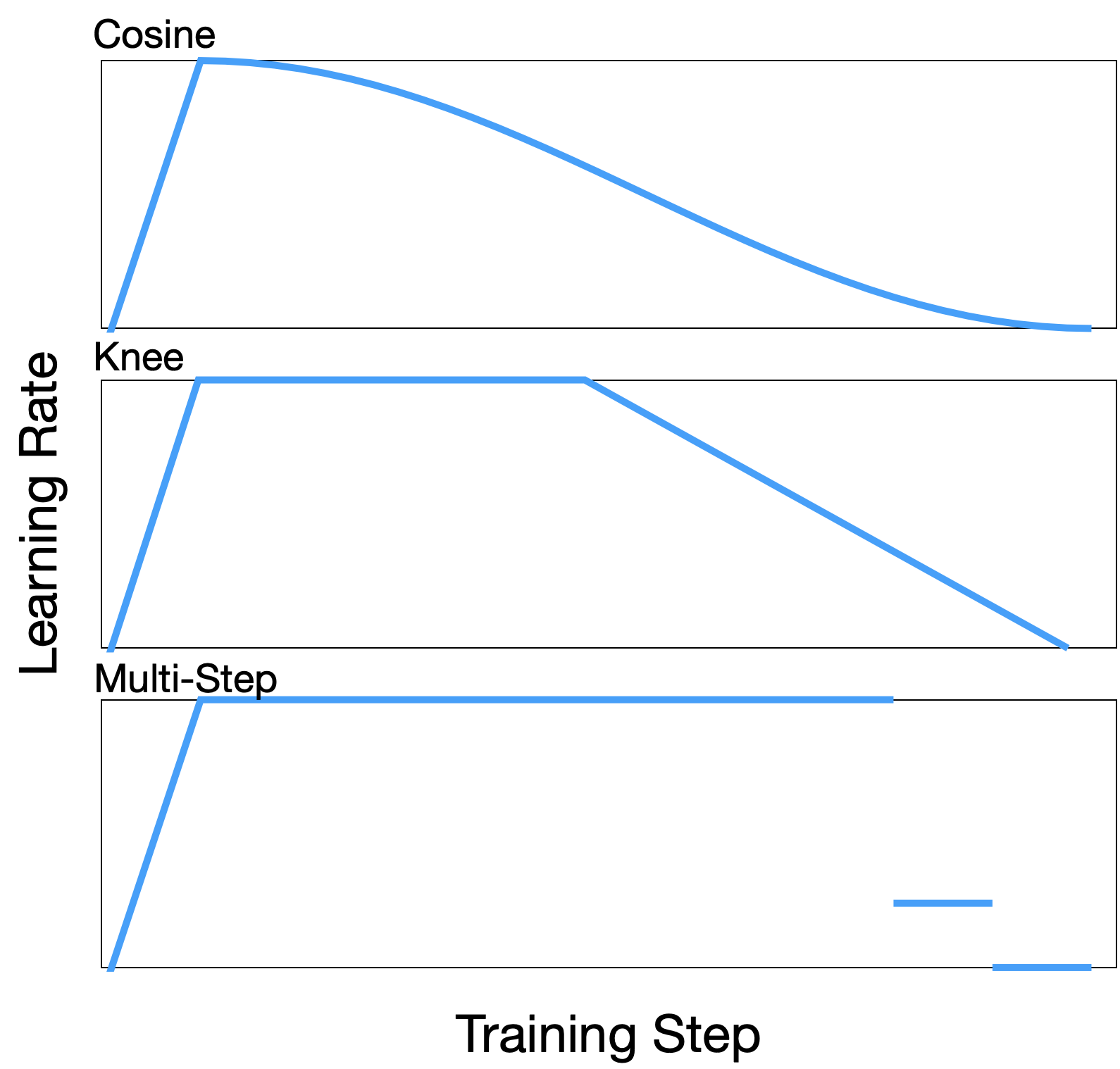}
        \caption{The learning rate curves of cosine~\cite{cos}, Knee~\cite{Knee}, and multi-step~\cite{deepseek} learning rate schedules.}
        \label{fig:diff-lr-curve_simple}
    \end{center}
\end{figure}

\begin{table}[t]
    \centering
    \begin{tabular}{l c c ccc}
        \toprule
        \multirow{2}{*}{\bf LRS} & \multirow{2}{*}{\bf TP} & \multirow{2}{*}{\bf Cost} & \multicolumn{3}{c}{\bf PPL} \\
        \cmidrule(lr){4-6}
         & & & \bf V2 & \bf V3 & \bf V4 \\
        \midrule
        \multirow{3}{*}{Cos} & PTFS       & 1.00$\times$ & 20.84 & 19.28 & 18.36 \\
        & CPT                             & 0.40$\times$ & 21.11 & 19.70 & 18.87  \\
        \cdashline{2-6}
        & $\Delta$                        & -            & -0.27 & -0.42 & -0.51 \\
        \midrule
        \multirow{3}{*}{Knee} & PTFS      & 1.00$\times$ & 20.22 & 18.80 & 17.98 \\
        & CPT                             & 0.40$\times$ & 20.56 & 19.27 & 18.52 \\
        \cdashline{2-6}
        & $\Delta$                        & -            & -0.34 & -0.47 & -0.54 \\
        \midrule
        \multirow{3}{*}{Multi} & PTFS     & 1.00$\times$ & 20.28 & 18.88 & 18.06 \\
        & CPT                             & 0.40$\times$ & 20.62 & 19.37 & 18.65 \\
        \cdashline{2-6}
        & $\Delta$                        & -            & -0.34 & -0.49 & -0.59 \\
        \bottomrule
    \end{tabular}
    \caption{The comparison between PTFS and CPT for training four versions of LLMs. ``LRS'' and ``TP'' indicate learning rate schedule and training paradigm, respectively. ``V*'' means the *-th version of LLM. Notably, regardless of PTFS or CPT, the learning rate curve and pre-training performance of the first version remain identical. Thus, we do not report the performance of the first version in all experiments.}
    \label{tab:paradigm-gap-1.2B}
\end{table}

\paragraph{Learning Rate Schedule} 
We conduct experiments with three learning rate schedules: cosine~\cite{cos}, Knee~\cite{Knee}, and multi-step~\cite{deepseek} learning rate schedules.\footnote{We also evaluate constant and inverse square root learning rate schedules, both of which yield inferior performance compared to the three selected schedules.}
The specific learning rate curves of these schedules are plotted in Figure~\ref{fig:diff-lr-curve_simple}.
Notably, cosine learning rate schedule is the most commonly used one for training LLMs~\cite{llm-survey1}, and both Knee and multi-step learning rate schedules can achieve comparable or even superior performance than cosine learning rate schedule.
For all learning rate schedules, we implement a linear warm-up phase of 2K steps (approximately 2.1B tokens).
Besides, we set the maximum and minimum learning rates for these schedules to 3e-4 and 3e-5, respectively.

\paragraph{Dataset} 
Similar to LLaMA~\cite{llama,llama2}, our training corpus comprises a mixture of data from publicly available sources, including code, paper, Wikipedia, books, mathematics, CommonCrawl and C4, webpage, translation and others.
In total, our training data contains 764M English and Chinese samples.
Due to the limitation of GPU resource, we do not experiment with the entire dataset. 
To simulate the scenario of version updates, we perform non-replacement sampling on the training data to obtain 10.5B tokens as the newly-added data for each update.
Hence, when using PTFS, we train four versions of LLMs from scratch with 10.5B, 21B, 31.5B, and 42B tokens, respectively. 
By contrast, using CPT to update the LLMs only involves the newly-added 10.5B tokens each time.

\paragraph{Evaluation} Following previous studies~\cite{ELLE, rewarm, deepseek}, we mainly use perplexity (PPL) to evaluate the pre-training performance of LLMs. 
Meanwhile, we also focus on the training cost of each paradigm, defined as the total training steps required for different versions.

\subsection{Comparison Between PTFS and CPT}

Experimental results are shown in Table~\ref{tab:paradigm-gap-1.2B}.
It is evident that CPT has lower training cost, whereas PTFS achieves superior performance.
More importantly, as the version updates progress, the performance gap between PTFS and CPT progressively widens.

To understand the underlying cause of this phenomenon, we focus on the learning rate, the key distinction between PTFS and CPT during version updates of LLMs. 
Using the cosine learning rate schedule, we conduct two groups of experiments to examine its impact on updated LLM performance across the two stages of CPT: 
1) preparing an initialization checkpoint, and 
2) continual pre-training based on this checkpoint.

\begin{figure}[t]
    \begin{center}
        \includegraphics[height=0.21\textheight]{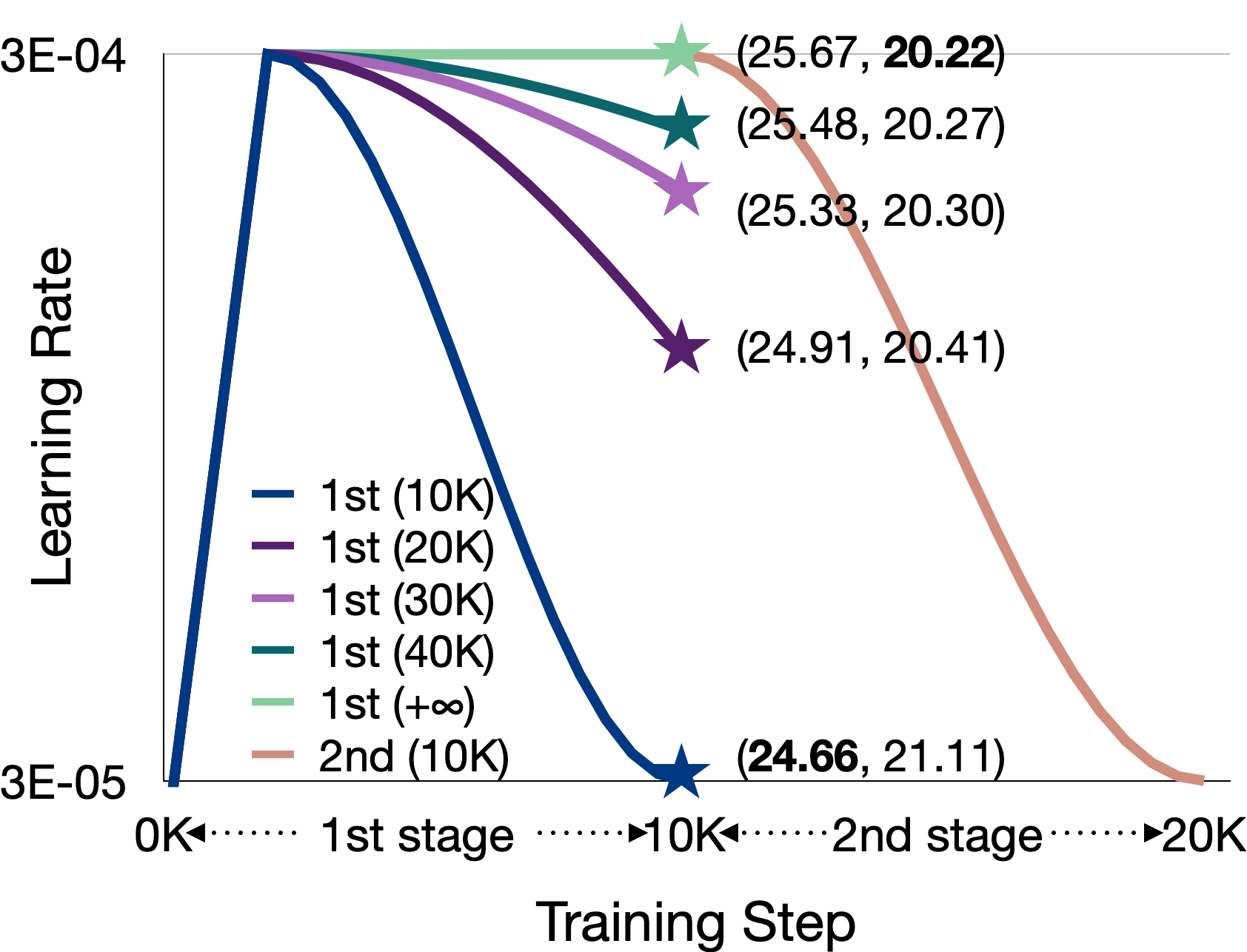}
        \caption{The effect of learning rate adjustment in the first stage. In the first stage, we vary the cosine cycle length as 10K, 20K, 30K, 40K and +$\infty$ steps, respectively, where the checkpoints at the 10K-th steps are selected as the initialization ones for the subsequent 10K-steps continual pre-training. ``($\cdot$,$\cdot$)'' indicates the PPLs of the initialization checkpoint and corresponding updated LLM.}
        \label{fig:first-stage}
    \end{center}
\end{figure}

\begin{figure}[t]
    \begin{center}
        \includegraphics[height=0.21\textheight]{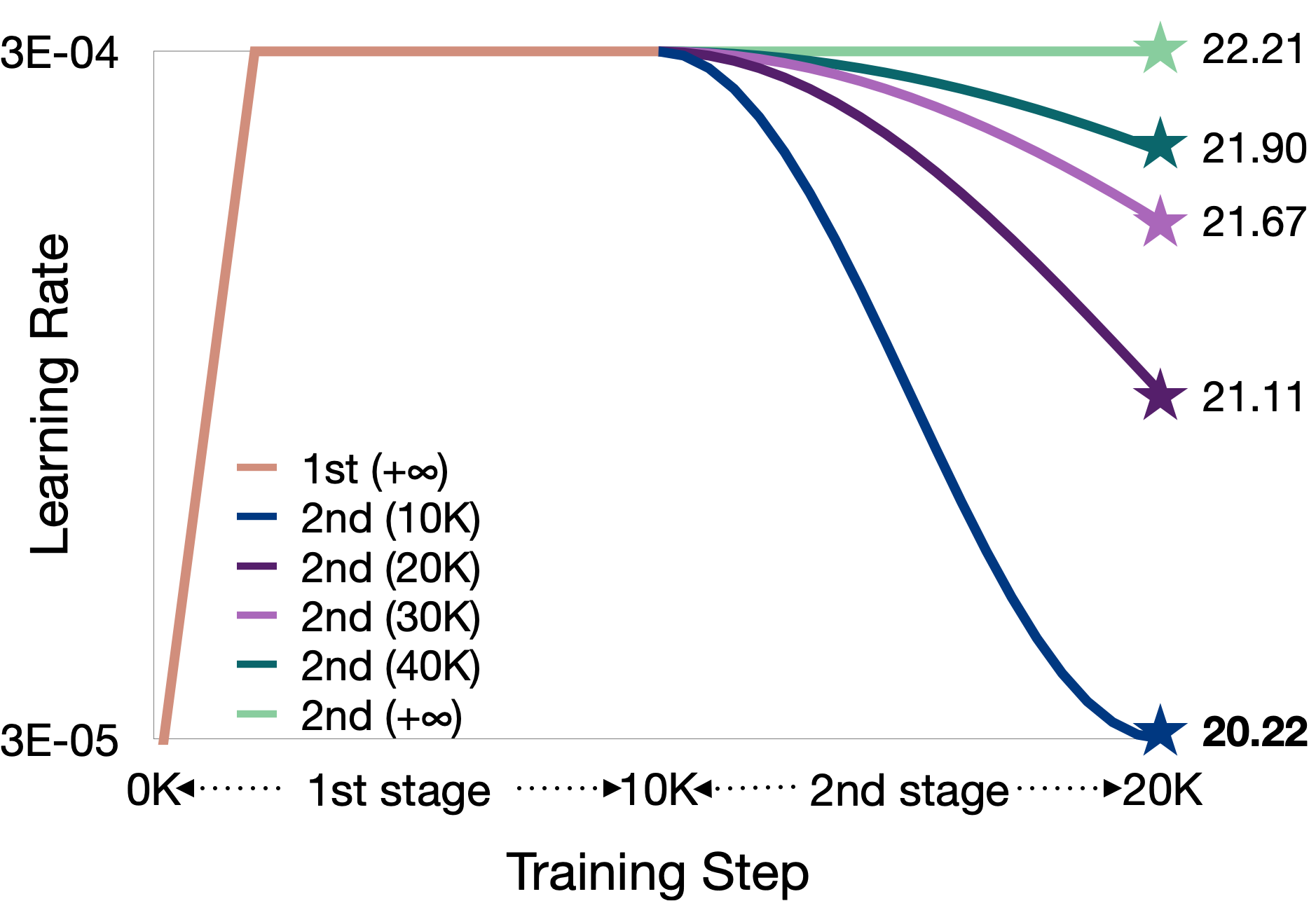}
        \caption{The effect of learning rate adjustment in the second stage. In the first stage, we directly use the maximal learning rate after warm-up. During the second stage, we try cosine cycle length with 10K, 20K, 30K, 40K and +$\infty$ steps, respectively, where the PPLs of LLMs at the 20K-th steps are compared.}
        \label{fig:second-stage}
    \end{center}
\end{figure}

\paragraph{Effect of Learning Rate Adjustment During the First Stage}
As depicted in Figure~\ref{fig:first-stage}, in the first group of experiments, we vary the cosine cycle length across 10K, 20K, 30K, 40K, and $+\infty$ steps, respectively.
The checkpoints at the 10K-th steps are selected as initialization checkpoints for the second stage.
Then, we continually pre-train LLMs for 10K steps, where the learning rate gradually decays from its maximum to minimum.
Referring back to Figure~\ref{fig:first-stage}, we observe that {\bf with the increase in the cosine cycle length during the first stage, the performance of the initialization checkpoint drops, whereas its corresponding updated LLM performs better}.
Therefore, we conclude that a large learning rate in the first stage benefits continual pre-training in the second stage.

\paragraph{Effect of Learning Rate Adjustment During the Second Stage}
Based on the above conclusion, we directly set the cosine cycle length as +$\infty$ steps in the first stage, as illustrated in Figure~\ref{fig:second-stage}.
Then, during continual pre-training, we experiment with the cosine learning rate schedule using different cosine cycle lengths: 10K, 20K, 30K, 40K, +$\infty$ steps, and report the performance of updated LLMs at the 20K-th steps.
As shown in Figure~\ref{fig:second-stage}, it is evident that {\bf a complete learning rate decay process enables the updated LLMs to achieve the best performance}.
This finding is consistent with the results from the first group of experiments mentioned above.
In other words, when the learning rate undergoes complete decay during the first stage, the performance of the initialization checkpoint is also optimal.

Based on the findings of the above two groups of experiments, we conclude that CPT is difficult to achieve good performance across different versions of LLMs.
Specifically, according to the findings from the second group of experiments, if the current LLM is expected to achieve optimal performance, its learning rate in the second stage should undergo a complete decay process. 
However, such decay results in a lower learning rate in the first stage of the subsequent update, further degrading the performance of the updated LLM.

\section{Our Paradigm}
\label{sec:ourparadigm}

Based on the conclusions from Section~\ref{sec:preliminary}, we propose a learning rate path switching paradigm for version updates of LLMs in this section.
The training cost of our paradigm is lower than that of PTFS, and it achieves significantly better performance than CPT, with performance even comparable to that of PTFS.

\begin{table}[t]
    \centering
    \begin{tabular}{ccc ccc}
        \toprule
        \multirow{2}{*}{\bf LRS} & \multirow{2}{*}{\bm{$\alpha$}} & \multirow{2}{*}{\bf Cost} & \multicolumn{3}{c}{\bf PPL} \\
        \cmidrule(lr){4-6}
         & & &\bf V2 & \bf V3 & \bf V4 \\
        \midrule
        \multirow{4}{*}{Cos}   & 0.2 & \bf 0.49$\times$ & 20.34 & 19.13 & 18.44 \\
                               & 0.4 & 0.53$\times$     & 20.16 & 18.91 & 18.21 \\
                               & 0.6 & 0.58$\times$     & \bf 20.13 & 18.81 & 18.09 \\
                               & 0.8 & 0.62$\times$     & 20.15 & \bf 18.77 & \bf 18.02 \\
        \midrule
        \multirow{4}{*}{Knee}  & 0.2 & \bf 0.49$\times$ & 20.33 & 19.12 & 18.42 \\
                               & 0.4 & 0.53$\times$     & 20.16 & 18.91 & 18.20 \\
                               & 0.6 & 0.58$\times$     & \bf 20.12 & 18.81 & 18.08 \\
                               & 0.8 & 0.62$\times$     & 20.15 & \bf 18.77 & \bf 18.01 \\
        \midrule
        \multirow{4}{*}{Multi} & 0.2 & \bf 0.49$\times$ & 20.33 & 19.08 & 18.37 \\
                               & 0.4 & 0.53$\times$     & \bf 20.29 & 18.91 & 18.16 \\
                               & 0.6 & 0.58$\times$     & 20.40 & \bf 18.88 & 18.09 \\
                               & 0.8 & 0.62$\times$     & 20.63 & 18.91 & \bf 18.06 \\
        \bottomrule
    \end{tabular}
    \caption{The effect of hyper-parameter $\alpha$ on the pre-training performance and training cost of our paradigm. Experiments are conducted on LLaMA-1.2B.}
    \label{tab:diff-fast-decay}
\end{table}

\begin{table}[t]
    \centering
    \scalebox{0.98}{
    \begin{tabular}{l c c ccc}
        \toprule
        \multirow{2}{*}{\bf LRS} & \multirow{2}{*}{\bf TP} & \multirow{2}{*}{\bf Cost} & \multicolumn{3}{c}{\bf PPL} \\
        \cmidrule(lr){4-6}
         & & & \bf V2 & \bf V3 & \bf V4 \\
        \midrule
        \multirow{3}{*}{Cos} & PTFS & 1.00$\times$ & 20.84 & 19.28 & 18.36 \\
        & CPT & \bf 0.40$\times$ & 21.11 & 19.70 & 18.87  \\
        \cdashline{2-6}
        & Ours & 0.58$\times$ & \bf 20.13 & \bf 18.81 & \bf 18.09 \\
        \midrule
        \multirow{3}{*}{Knee} & PTFS & 1.00$\times$ & 20.22 & \bf 18.80 & \bf 17.98 \\
        & CPT & \bf 0.40$\times$ & 20.56 & 19.27 & 18.52 \\
        \cdashline{2-6}
        & Ours & 0.58$\times$ & \bf 20.12 & 18.81 & 18.08 \\
        \midrule
        \multirow{3}{*}{Multi} & PTFS & 1.00$\times$ & \bf 20.28 & \bf 18.88 & \bf 18.06 \\
        & CPT & \bf 0.40$\times$ & 20.62 & 19.37 & 18.65 \\
        \cdashline{2-6}
        & Ours & 0.58$\times$ & 20.40 & \bf 18.88 & 18.09 \\
        \bottomrule
    \end{tabular}}
    \caption{The comparison of different paradigms for training four versions of LLaMA-1.2B.}
    \label{tab:diff-paradigm-1.2B}
\end{table}

\subsection{Paradigm Overview}

Let us revisit Figure~\ref{fig:diff-paradigm}, which shows the learning rate curves of our paradigm applied to the cosine learning rate schedule. 
Please note that our paradigm is also applicable to other schedules, such as Knee and multi-step and so on.
The learning rate curve of our paradigm comprises one main path and multiple branching paths, each of which corresponds to one version update.
On the main path, we pre-train the LLM from scratch with the maximal learning rate, providing initialization checkpoints for subsequent version updates.
When we want to obtain an updated LLM, we directly use the current checkpoint of the main path as the initialization one, and then perform continual pre-training. 
During this process, the learning rate undergoes a complete fast-decaying process, effectively ensuring the performance of the updated LLM.
Meanwhile, on the main path, we still use newly-added data to pre-train the existing checkpoint with the maximal learning rate, so as to facilitate subsequent updates.

Obviously, our paradigm has lower training cost than PTFS, as it conducts continual pre-training based on the initialization checkpoints from the main path.
Unlike CPT, these checkpoints are obtained through training from scratch with the maximum learning rate, which enables the updated LLMs to achieve better performance, as analyzed in Section~\ref{sec:preliminary}.
The following experiments fully confirm the superiority of our paradigm in balancing model performance and training cost.

\begin{table*}[t]
    \centering
    \scalebox{0.96}{
    \begin{tabular}{cc c rr rrr rrr r}
        \toprule
        \bf Ver. & \bf TP & {\small C\textsuperscript{3}} & {\small GSM8K} & {\small MMLU} & {\small CSL} & {\small C-EVAL} & {\small BBH} & {\small CMMLU} & {\small GAOKAO} & {\small AGIEval} & \bf AVG \\
        \midrule
        \multirow{3}{*}{V2} & PTFS & 38.00 & 4.63 & \bf 24.00 & 38.25 & \bf 30.09 & 17.43 & 25.37 & 18.10 & 14.59 & 23.38 \\
         & CPT  & 37.00 & 4.09 & 23.52 & 35.11 & 27.42 & 18.55 & \bf 25.63 & \bf 18.86 & 13.40 & 22.62 \\
        \cdashline{2-12}
         & Ours & \bf 38.60 & \bf 5.08 & 22.94 & \bf 39.08 & 28.38 & \bf 20.79 & 24.88 & 18.48 & \bf 14.73 & \bf 23.66 \\
        \midrule
        \multirow{3}{*}{V3} & PTFS & 40.30 & 3.34 & \bf 24.33 & \bf 39.17 & 25.85 & 17.11 & \bf 25.30 & \bf 22.03 & 14.34 & 23.53 \\
         & CPT  & 38.30 & \bf 4.70 & 23.32 & 36.40 & 28.38 & \bf 21.11 & 24.76 & 17.85 & 13.47 & 23.14 \\
        \cdashline{2-12}
         & Ours & \bf 42.10 & 4.63 & 23.22 & 34.91 & \bf 29.35 & 19.70 & 24.73 & 19.24 & \bf 14.90 & \bf 23.64 \\
        \midrule
        \multirow{3}{*}{V4} & PTFS & 35.70 & 4.25 & \bf 24.93 & 38.75 & 27.04 & 16.73 & \bf 24.97 & \bf 21.01 & 14.10 & 23.05 \\
         & CPT  & \bf 43.90 & 4.55 & 22.20 & 38.69 & 27.19 & 21.62 & 24.43 & 18.23 & 13.50 & 23.81 \\
        \cdashline{2-12}
         & Ours & 41.90 & \bf 5.53 & 24.09 & \bf 40.24 & \bf 27.71 & \bf 21.84 & 24.78 & 17.24 & \bf 14.40 & \bf 24.19 \\
        \bottomrule
    \end{tabular}}
    \caption{The performance of LLMs across different versions on downstream tasks. ``Ver.'' indicates the version number of the LLMs. Additional experimental results for LLMs with larger model sizes or data sizes are listed in Appendix~\ref{sec:downstream}.}
    \label{tab:down-task}
\end{table*}

\subsection{Time Complexity Analysis}
\label{subsec:complexity-analysis}

To further compare different training paradigms in terms of training cost, we define their time complexity functions as the total training steps of version updates.

Before providing our definitions, we first introduce two symbols to facilitate the subsequent descriptions:
1) $N_v$: the number of version updates of LLMs;
2) $T$: the amount of data added for each update, assuming it remains consistent.
When updating the $i-th$ version of LLMs, PTFS requires updating $i T (1\leq{i}\leq{N_v})$ steps each time, CPT needs to train for $T$ steps, and our paradigm requires training $T + \alpha T$ steps, where $\alpha~(0\leq\alpha\leq1)$ controls the proportion of fast-decaying steps to the total steps in each update.

Formally, the time complexity functions of PTFS, CPT and our paradigm can be described as follows:
\begin{equation}
\begin{split}
    \mathbf{C}_{\text{ptfs}}(N_v) &= \sum_{i=1}^{N_v} i T = 0.5TN_v^2 + 0.5TN_v, \\
    \mathbf{C}_{\text{cpt}}(N_v) &= \sum_{i=1}^{N_v} T = TN_v, \\
    \mathbf{C}_{\text{ours}}(N_v) &= \sum_{i=1}^{N_{v}-1} (T + {\alpha}T) + T \\
    &= (1 + \alpha)TN_v - \alpha{T}.\nonumber
\end{split}
\end{equation}
Please note that, for the last version, the additional main path training for preparing the initialization checkpoint for the next update can be omitted, which counts as $\alpha T$ steps. Thus, only $T$ steps are required.

Comparing the above functions, we observe that $\mathbf{C}_{\text{ptfs}}(N_v)$ is a quadratic function in terms of $N_v$, whereas both $\mathbf{C}_{\text{cpt}}(N_v)$ and $\mathbf{C}_{\text{ours}}(N_v)$ are linear functions.
Moreover, the gaps between $\mathbf{C}_{\text{ptfs}}(N_v)$ and the other two functions significantly widens as $N_v$ increases.
For example, when $N_v = 4$, the values of these three time complexity functions are $10T$, $4T$ and $5.8T$, respectively.
When $N_v = 10$, the gaps widen as the values of these functions increase to $55T$, $10T$ and $15.4T$.

\section{Experiment}

In this section, we still use the settings of the preliminary study to conduct more experiments, comparing the performance and training cost of different training paradigms.

\subsection{Effect of Hyper-Parameter $\alpha$}

As described in Section~\ref{sec:ourparadigm}, $\alpha$ is one of the most important hyper-parameters in our paradigm, as it controls the proportion of fast-decaying steps to the total steps in each update.
The fast-decaying steps influence model performance and training cost of our paradigm.
To select an optimal ${\alpha}$ value, we experiment with different $\alpha$ values, ranging from 0.2 to 0.8 with an interval of 0.2, and then observe the changes in pre-training performance and training cost.

Experimental results are listed in Table~\ref{tab:diff-fast-decay}, showing that the overall performance of LLMs across different versions is optimal at $\alpha=0.6$ and $\alpha=0.8$. 
However, when $\alpha=0.6$, our paradigm achieves lower training cost. 
Thus, we adopt $\alpha=0.6$ in subsequent experiments.

\subsection{Main Experiments}

Then, we compare different paradigms in terms of training cost, pre-training performance and downstream performance.
To comprehensively examine our paradigm, we conduct a series of experiments with the three aforementioned learning rate schedules.

\paragraph{Pre-Training Performance} 
From Table~\ref{tab:diff-paradigm-1.2B}, we observe that, {\bf compared to PTFS, our paradigm reduces the total training cost to 58\% while maintaining comparable pre-training performance}.
Particularly, when using the cosine learning rate schedule, our paradigm even slightly outperform PTFS. 
On the other hand, as expected, the training cost of our paradigm is still higher than that of CPT, however, it always achieves better performance than CPT, regardless of the schedule used.
Overall, our paradigm achieves a better balance between pre-training performance and total training cost during version updates of LLMs.\footnote{We also compare our paradigm with CPT based on equal training cost, with results detailed in Appendix \ref{sec:cpt_ours}. Besides, we also compare PTFS, CPT and our paradigm in the scenario with varying data increments. The corresponding results are listed in Appendix \ref{sec:inconsistent_data}.}

\paragraph{Performance on Downstream Tasks}
Furthermore, we investigate the performance of different training paradigms across nine downstream tasks, including C\textsuperscript{3}~\cite{c3}, GSM8K~\cite{GSM8K}, MMLU~\cite{MMLU}, CSL~\cite{csl}, C-EVAL~\cite{ceval},  BBH~\cite{BBH}, CMMLU~\cite{CMMLU}, GAOKAO~\cite{GAOKAO} and AGIEval~\cite{AGIEval}.
To this end, we first construct a general supervised fine-tuning (SFT) dataset with 1.8B tokens and then we perform SFT on each of the four versions of the updated LLMs. 

From the results listed in Table~\ref{tab:down-task}, we clearly find that our paradigm can still obtain better average performance than PTFS and CPT, which further proves the effectiveness of our paradigm.

\subsection{Generalization of Our Paradigm}
\label{subsec:generalization}

Subsequently, we explore the generalization of our paradigm in the following aspects, including model architecture, model size, data size, and maximum learning rate, all of which are crucial for the practical applications of LLMs.
In all of these experiments, we maintain the use of the cosine learning rate schedule.

\begin{table}[t]
    \centering
    \scalebox{0.98}{
    \begin{tabular}{l c c ccc}
        \toprule
        \multirow{2}{*}{\bf LRS} & \multirow{2}{*}{\bf TP} & \multirow{2}{*}{\bf Cost} & \multicolumn{3}{c}{\bf PPL} \\
        \cmidrule(lr){4-6}
         & & & \bf V2 & \bf V3 & \bf V4 \\
        \midrule
        \multirow{3}{*}{Cos} & PTFS & 1.00$\times$ & 20.94 & 19.35 & 18.41 \\
        & CPT & \bf 0.40$\times$ & 21.23 & 19.78 & 18.92 \\
        \cdashline{2-6}
        & Ours & 0.58$\times$ & \bf 20.23 & \bf 18.87 & \bf 18.11 \\
        \midrule
        \multirow{3}{*}{Knee} & PTFS & 1.00$\times$ & 20.30 & \bf 18.84 & \bf 17.98 \\
        & CPT & \bf 0.40$\times$ & 20.67 & 19.34 & 18.56 \\
        \cdashline{2-6}
        & Ours & 0.58$\times$ & \bf 20.20 & 18.85 & 18.09 \\
        \midrule
        \multirow{3}{*}{Multi} & PTFS & 1.00$\times$ & \bf 20.37 & \bf 18.92 & \bf 18.06 \\
        & CPT & \bf 0.40$\times$ & 20.74 & 19.44 & 18.68 \\
        \cdashline{2-6}
        & Ours & 0.58$\times$ & 20.49 & \bf 18.92 & 18.09 \\
        \bottomrule
    \end{tabular}}
    \caption{The generalization of our paradigm in terms of model architecture. Based on Qwen-1.2B, we conduct experiments with the same setting as LLaMA-1.2B.}
    \label{tab:diff-paradigm-1.2B_qwen}
\end{table}

\begin{table}[t]
    \centering
    \begin{tabular}{l c ccc}
        \toprule
        \multirow{2}{*}{\bf Size} & \multirow{2}{*}{\bf TP} & \multicolumn{3}{c}{\bf PPL} \\
        \cmidrule(lr){3-5}
         & & \bf V2 & \bf V3 & \bf V4 \\
        \midrule
        \multirow{3}{*}{203M} & PTFS & 30.97 & 29.50 & 28.65 \\
         & CPT  & 31.31 & 29.90 & 29.07\\
        \cdashline{2-5}
         & Ours & \bf 30.25 & \bf 28.94 & \bf 28.19\\
        \midrule

        \multirow{3}{*}{406M} & PTFS & 26.58 & 25.06 & 24.19 \\
         & CPT  & 26.89 & 25.49 & 24.67 \\
        \cdashline{2-5}
         & Ours & \bf 25.85 & \bf 24.52 & \bf 23.79 \\
        \midrule

        \multirow{3}{*}{608M} & PTFS & 23.12 & 21.75 & 20.93 \\
         & CPT  & 23.50 & 22.26 & 21.52 \\
        \cdashline{2-5}
         & Ours & \bf 22.59 & \bf 21.43 & \bf 20.77 \\
        \midrule
        
        \multirow{3}{*}{1.2B} & PTFS & 20.84 & 19.28 & 18.36 \\
         & CPT  & 21.22 & 19.79 & 18.97 \\
        \cdashline{2-5}
         & Ours & \bf 20.13 & \bf 18.81 & \bf 18.09 \\
        \midrule

        \multirow{3}{*}{2.1B} & PTFS & 18.33 & 16.88 & 16.04 \\
         & CPT  & 18.76 & 17.47 & 16.72 \\
        \cdashline{2-5}
         & Ours & \bf 17.82 & \bf 16.63 & \bf 15.97 \\
        \midrule

        \multirow{3}{*}{3.1B} & PTFS & 17.22 & 15.87 & \bf 15.07 \\
         & CPT  & 17.67 & 16.48 & 15.77 \\
        \cdashline{2-5}
         & Ours & \bf 16.84 & \bf 15.72 & 15.09 \\
         
        \bottomrule
    \end{tabular}
    \caption{The generalization of our paradigm in terms of model size. The model sizes range from 203M to 3.1B.}
    \label{tab:different-size}
\end{table}

\paragraph{Model Architecture}
To demonstrate the generalization of our paradigm on model architecture, we use Qwen-1.2B~\cite{qwen} to re-conduct experiments with the same setting as LLaMA-1.2B.

Similar to the experimental results of LLaMA-1.2B presented in Table~\ref{tab:diff-paradigm-1.2B}, the experimental results of Qwen-1.2B shown in Table~\ref{tab:diff-paradigm-1.2B_qwen} further demonstrate the superiority of our paradigm in balancing model performance and training cost. 
This validates the generalization of our paradigm in terms of model architecture.

\paragraph{Model Size}
We then focus on the generalization of our paradigm on model size.
To this end, we vary the number of model parameters to conduct experiments.
In total, we consider the following six model sizes: 203M, 406M, 608M, 1.2B, 2.1B, 3.1B, of which detailed hyper-parameters are listed in Appendix~\ref{sec:training-detail}.

From the results shown in Table~\ref{tab:different-size}, we observe that our paradigm achieves pre-training performance comparable to PTFS across different sizes of LLMs and outperforms CPT.
This validates the generalization of our paradigm in terms of model size.

\paragraph{Data Size}
Next, we switch our attention to the generalization of our paradigm on data size.
To do this, we conduct experiments using different sizes of training data: 21B, 42B, and 168B tokens.
Correspondingly, the training steps are 5K, 10K and 40K for each LLM update, respectively.

As shown in Table~\ref{tab:1.2B-160K}, our paradigm achieves optimal pre-training performance across different data sizes, which further demonstrates the generalization of our paradigm.

\begin{table}[t]
    \centering
    \begin{tabular}{c c ccc}
        \toprule
        \bf \multirow{2}{*}{\bf Data} & \multirow{2}{*}{\bf TP} & \multicolumn{3}{c}{\bf PPL} \\
        \cmidrule(lr){3-5}
         & & \bf V2 & \bf V3 & \bf V4 \\
         \midrule
         \multirow{3}{*}{21B}  & PTFS & 24.66 & 22.31 & 20.84 \\
                               & CPT  & 25.10 & 22.84 & 21.56 \\
         \cdashline{2-5}
                               & Ours  & \bf 23.59 & \bf 21.41 & \bf 20.27 \\
         \midrule
         \multirow{3}{*}{42B}  & PTFS & 20.84 & 19.28 & 18.36 \\
                               & CPT  & 21.11 & 19.70 & 18.87 \\
         \cdashline{2-5}
                               & Ours  & \bf 20.13 & \bf 18.81 & \bf 18.09 \\
         \midrule
         \multirow{3}{*}{168B} & PTFS & 16.70 & 15.97 & 15.54 \\
                               & CPT  & 16.90 & 16.25 & 15.86 \\
         \cdashline{2-5}
                               & Ours  & \bf 16.47 & \bf 15.86 & \bf 15.51 \\
        \bottomrule
    \end{tabular}
    \caption{The generalization of our paradigm in terms of data size. The total data sizes (for four versions) range from 21B to 168B.}
    \label{tab:1.2B-160K}
\end{table}

\begin{table}[t]
    \centering
    \begin{tabular}{l c ccc}
        \toprule
        \multirow{2}{*}{\bf MLR} & \multirow{2}{*}{\bf TP} & \multicolumn{3}{c}{\bf PPL} \\
        \cmidrule(lr){3-5}
         & & \bf V2 & \bf V3 & \bf V4 \\
        \midrule
        \multirow{3}{*}{5e-5} & PTFS & 34.78 & 29.53 & 26.65 \\
         & CPT  & 35.23 & 30.08 & 27.23 \\
        \cdashline{2-5}
         & Ours & \bf 29.99 & \bf 25.54 & \bf 23.27 \\
        \midrule

        \multirow{3}{*}{1e-4} & PTFS & 26.34 & 23.28 & 21.57 \\
         & CPT  & 26.64 & 23.70 & 22.04 \\
        \cdashline{2-5}
         & Ours & \bf 23.89 & \bf 21.32 & \bf 19.97 \\
        \midrule
        
        \multirow{3}{*}{3e-4} &PTFS & 20.84 & 19.28 & 18.36 \\
         & CPT  & 21.22 & 19.79 & 18.97  \\
        \cdashline{2-5}
         & Ours & \bf 20.13 & \bf 18.81 & \bf 18.09 \\
        \midrule

        \multirow{3}{*}{5e-4} & PTFS & 19.89 & 18.62 & \bf 17.85 \\
         & CPT  & 20.17 & 19.05 & 18.38 \\
        \cdashline{2-5}
         & Ours & \bf 19.53 & \bf 18.45 & \bf 17.85 \\
        \midrule

        \multirow{3}{*}{8e-4} & PTFS & 19.38 & \bf 18.26 & \bf 17.58 \\
         & CPT  & 19.69 & 18.73 & 18.16 \\
        \cdashline{2-5}
         & Ours & \bf 19.22 & 18.30 & 17.78 \\
        
        \bottomrule
    \end{tabular}
    \caption{The generalization of our paradigm in terms of the maximum learning rate. The maximum learning rate ranges from 5e-5 to 8e-4. ``MLR'' indicates the maximum learning rate.}
    \label{tab:different-lr}
\end{table}

\paragraph{Maximum Learning Rate}
Finally, we aim to verify the generalization of our paradigm in terms of the maximum learning rate.
We conduct experiments by setting the maximum learning rates as 5e-5, 1e-4, 3e-4, 5e-4, 8e-4, respectively. 

As shown in Table~\ref{tab:different-lr}, as the maximum learning rate increases, our paradigm consistently achieves better or comparable performance than PTFS, and significantly outperforms CPT.
This strongly highlights the generalization of our paradigm in terms of the maximum learning rate.

\section{Related Work}

\paragraph{Continual Training}
As one of the most direct approaches for version updates of LLMs, continual training has attracted increasing attention, of which related studies can be broadly categorized into the following four types:
1) methods introducing additional parameters~\cite{ke2022continual, ke2023continual, song2023conpet, PENG2024Scalable}, 2) prompt-based methods~\cite{wang2022learning, wang2022dualprompt, razdaibiedina2023progressive}, 3) multi-stage training methods~\cite{DBLP:conf/acl/LiuYLZLZZS20, DBLP:conf/acl/ZhouMZZWS22, DBLP:journals/pami/ZhouLMZXWZS23, DBLP:conf/emnlp/LiuHYLSH23, DBLP:conf/acl/HuangCWYLSYS24}, and 4) scenario-specific methods~\cite{peng2023semiparametric, gogoulou2023study, xie2023efficient}.
Significantly different from the above studies, our paradigm comprises one main learning rate path, where we perform pre-training from scratch with the maximal learning rate, and multiple learning rate branching paths, where we perform continual pre-training with a complete learning rate decay process.
Thus, our paradigm achieve a better trade-off between the performance and training cost than PTFS and CPT.

\paragraph{Learning Rate} 
The learning rate is one of the most crucial hyper-parameters for training LLMs.
Existing learning rate schedules can be broadly divided into the following four policies~\cite{lr1,lr2,rethink-lr}:
1) Fixed learning rate policy, such as constant learning rate schedule; 
2) Decaying learning rate policy, such as inverse square root learning rate schedule; 
3) Cyclic learning rate policy, such as cosine learning rate schedule; 
4) Composite learning rate policy, such as Knee and multi-step learning rate schedules. 
In addition, there are some recent studies exploring learning rate schedules for LLMs, including Warmup-Stable-Decay schedule~\cite{minicpm} and constant learning rate with cooldown~\cite{hf-scaling}.
Particularly, our paradigm is a well-designed training paradigm for version updates of LLMs, which is applicable to cosine, Knee, and multi-step and other learning rate schedules.

\section{Conclusion and Future Work}

This paper focuses on how to effectively balance model performance and training cost for version updates of LLMs. 
We begin by comparing two training paradigms: PTFS and CPT, concluding that PTFS achieves better pre-training performance, while CPT has lower training cost.
Through the analysis in the preliminary study, we find that 1) a large learning rate is beneficial for providing better initialization checkpoints for subsequent updates, and 2) a complete learning rate decay process enables the updated LLMs to achieve optimal performance.
Based on the above two findings, we propose a learning rate path switching paradigm for version updates of LLMs, which comprises one main path and multiple branching paths.
On the main path, we pre-train the LLMs with the maximal learning rate to provide superior initialization checkpoints for subsequent updates.
When an update is required, our paradigm switches from the main path to a branching path, undergoing a complete learning rate decay process.
Experimental results and further analyses strongly demonstrate the effectiveness and generalization of our paradigm.

In the future, we will further expand the practical scope of our paradigm.
Current research mainly focuses on the pre-training phase and does not include supervised fine-tuning, safety alignment, etc., which could be integrated into the fast-decaying stage of our paradigm.
Additionally, we plan to explore the applicability of our paradigm in the context of multimodal large language models.

\section*{Limitations}

Although the training cost of our paradigm is significantly lower than that of PTFS, it is still higher than that of CPT. 
Hence, we plan to design a precise method to determine the proportion of the fast-decaying steps to the total steps, which can further reduce the training cost of our paradigm.

\section*{Acknowledgements}
The project was supported by National Key R\&D Program of China (No. 2022ZD0160501), National Natural Science Foundation of China (No. 62276219), and the Public Technology Service Platform Project of Xiamen (No. 3502Z20231043).

\bibliography{custom}

\clearpage
\appendix

\begin{table}[t!]
    \centering
    \begin{tabular}{r r r r r}
        \toprule
        \bf Size & \bf MLR & \bf Hidden & \bf Head & \bf Layer \\
        \midrule
        203M & 1e-3 &   512 &  8 & 24 \\
        406M & 6e-4 & 1,024 & 16 & 12 \\
        608M & 6e-4 & 1,024 & 16 & 24 \\
        1.2B & 3e-4 & 1,536 & 16 & 24 \\
        2.1B & 3e-4 & 1,536 & 16 & 48 \\
        3.1B & 3e-4 & 8,192 & 32 & 40 \\
        \bottomrule
    \end{tabular}
    \caption{The detailed hyper-parameters of LLMs with different model sizes.}
    \label{tab:hyper-parameters}
\end{table}

\section{Detailed Hyper-Parameters}
\label{sec:training-detail}

In this work, we compare PTFS, CPT and our paradigm based on LLMs with different sizes, whose hyper-parameters are listed in Table~\ref{tab:hyper-parameters}.
Following \citeauthor{scaling4lm, gpt3}, we set smaller maximum learning rates for larger LLMs.
Besides, the minimum learning rate is configured to be 10\% of the maximum learning rate.

\begin{figure}[t]
    \begin{center}
        \includegraphics[height=0.28\textheight]{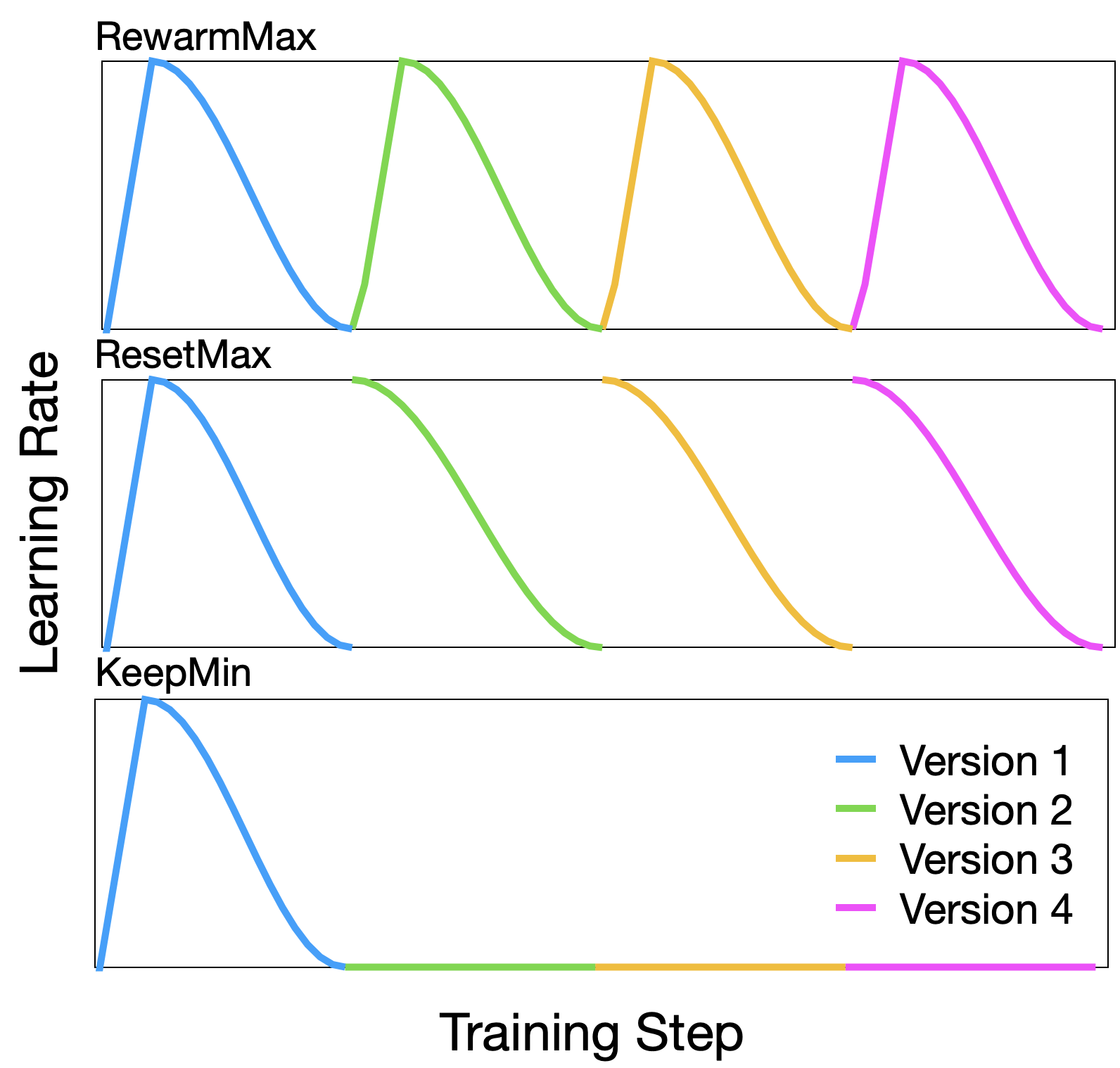}
        \caption{The learning rate curves of different adaptation method of CPT for version updates of LLMs. The learning rate curves are plotted based on cosine learning rate schedules.} 
        \label{fig:adaption-lr-curve}
    \end{center}
\end{figure}

\begin{table}
    \centering
    \begin{tabular}{l l ccc}
        \toprule
        \multirow{2}{*}{\bf LRS} & \multirow{2}{*}{\bf Variant} & \multicolumn{3}{c}{\bf PPL}  \\
        \cmidrule(lr){3-5}
         & & \bf V2 & \bf V3 & \bf V4 \\
        \midrule
        \multirow{3}{*}{Cos}   & RewarmMax & 21.22 & 19.79 & 18.97 \\
                               & ResetMax  & \bf 21.11 & \bf 19.70 & \bf 18.87 \\
                               & KeepMin   & 23.00 & 21.99 & 21.26 \\
        \midrule
        \multirow{3}{*}{Knee}  & RewarmMax & 20.74 & 19.46 & 18.70 \\
                               & ResetMax  & \bf 20.56 & \bf 19.27 & \bf 18.52 \\
                               & KeepMin   & 22.22 & 21.36 & 20.37 \\
        \midrule
        \multirow{3}{*}{Multi} & RewarmMax & 20.80 & 19.55 & 18.82\\
                               & ResetMax  & \bf 20.62 & \bf 19.37 & \bf 18.65 \\
                               & KeepMin   & 22.11 & 21.24 & 20.60 \\
        \bottomrule
    \end{tabular}
    \caption{The comparison among RewarmMax, ResetMax and KeepMin for CPT.}
    \label{tab:adaption-1.2b}
\end{table}

\section{CPT Variants}
\label{sec:opt-lr-set}

In order to adapt traditional CPT for version updates of LLMs, we compare three variants of CPT in Figure~\ref{fig:adaption-lr-curve}:
\begin{itemize}[leftmargin=*,topsep=0.2em,itemsep=0.2em,parsep=0.2em]
    \item {\bf RewarmMax}: Warm up the learning rate periodically, and use the learning rate schedule of the old version to train the new version of LLMs~\cite{rewarm}.
    
    \item {\bf ResetMax}: Directly set the learning rate as the maximum periodically, and use the learning rate schedule of the old version to train the new version of LLMs~\cite{rewarm}.

    \item {\bf KeepMin}: Keep the learning rate at the minimum by using a constant learning rate schedule to ensure the convergence of LLMs during training~\cite{gogoulou2023study}.
\end{itemize}
Experimental results are listed in Table~\ref{tab:adaption-1.2b}.
We observe that ResetMax achieves the best pre-training performance among these variants.
Therefore, we use ResetMax for the other experiments.

\section{Performance of Downstream Tasks}
\label{sec:downstream}

In addition to the standard training scale (LLaMA-1.2B trained for 42B tokens), we also evaluate LLMs with a larger training dataset (LLaMA-1.2B trained for 168B tokens) and a larger model size (LLaMA-3.1B trained for 42B tokens).
We report the performance of downstream tasks across different versions of LLMs, as shown in Table~\ref{tab:detailed-down-task}.
Experimental results show that our paradigm achieves superior average performance compared with PTFS and CPT across different training scales for downstream tasks.

\begin{table*}
    \centering
    \scalebox{0.89}{
    \begin{tabular}{ccc ccc ccc ccc r}
        \toprule
        \bf Scale & \bf Ver. & \bf TP & {\small C\textsuperscript{3}} & {\small GSM8K} & {\small MMLU} & {\small CSL} & {\small C-EVAL} & {\small BBH} & {\small CMMLU} & {\small GAOKAO} & {\small AGIEval} & \bf AVG \\
        \midrule
        \multirow{9}{*}{\parbox{0.7cm}{1.2B\\42B}} & \multirow{3}{*}{V2} & PTFS & 38.00 & 4.63 & \bf 24.00 & 38.25 & \bf 30.09 & 17.43 & 25.37 & 18.10 & 14.59 & 23.38 \\
         & & CPT  & 37.00 & 4.09 & 23.52 & 35.11 & 27.42 & 18.55 & \bf 25.63 & \bf 18.86 & 13.40 & 22.62 \\
        \cdashline{3-13}
         & & Ours & \bf 38.60 & \bf 5.08 & 22.94 & \bf 39.08 & 28.38 & \bf 20.79 & 24.88 & 18.48 & \bf 14.73 & \bf 23.66 \\
        \cmidrule(lr){2-13}
         & \multirow{3}{*}{V3} & PTFS & 40.30 & 3.34 & \bf 24.33 & \bf 39.17 & 25.85 & 17.11 & \bf 25.30 & \bf 22.03 & 14.34 & 23.53 \\
         & & CPT  & 38.30 & \bf 4.70 & 23.32 & 36.40 & 28.38 & \bf 21.11 & 24.76 & 17.85 & 13.47 & 23.14 \\
        \cdashline{3-13}
         & & Ours & \bf 42.10 & 4.63 & 23.22 & 34.91 & \bf 29.35 & 19.70 & 24.73 & 19.24 & \bf 14.90 & \bf 23.64 \\
        \cmidrule(lr){2-13}
         & \multirow{3}{*}{V4} & PTFS & 35.70 & 4.25 & \bf 24.93 & 38.75 & 27.04 & 16.73 & \bf 24.97 & \bf 21.01 & 14.10 & 23.05 \\
         & & CPT  & \bf 43.90 & 4.55 & 22.20 & 38.69 & 27.19 & 21.62 & 24.43 & 18.23 & 13.50 & 23.81 \\
        \cdashline{3-13}
         & & Ours & 41.90 & \bf 5.53 & 24.09 & \bf 40.24 & \bf 27.71 & \bf 21.84 & 24.78 & 17.24 & \bf 14.40 & \bf 24.19 \\
        \midrule
        \multirow{9}{*}{\parbox{0.7cm}{1.2B\\168B}} & \multirow{3}{*}{V2} & PTFS & 38.90 & 6.82 & 23.49 & 40.33 & \bf 29.27 & \bf 23.28 & 25.14 & \bf 23.29 & 14.39 & 24.99 \\
         & & CPT  & \bf 43.80 & 7.13 & 24.61 & 37.22 & 26.52 & 22.96 & 25.40 & 20.13 & 14.25 & 24.67 \\
        \cdashline{3-13}
         & & Ours & 43.20 & \bf 8.95 & \bf 25.43 & \bf 40.45 & 26.90 & 22.16 & \bf 25.45 & 18.73 & \bf 15.94 & \bf 25.25 \\
        \cmidrule(lr){2-13}
         & \multirow{3}{*}{V3} & PTFS & 47.40 & 8.49 & 25.04 & 42.42 & \bf 27.42 & \bf 26.88 & \bf 25.06 & 18.23 & 16.59 & \bf 26.39 \\
         & & CPT  & 40.30 & 8.42 & 24.30 & 41.61 & 26.30 & 24.07 & 24.59 & \bf 20.00 & \bf 18.00 & 25.29 \\
        \cdashline{3-13}
         & & Ours & \bf 47.70 & \bf 9.33 & \bf 25.35 & \bf 44.39 & 25.85 & 23.05 & 24.85 & 17.60 & 15.63 & 25.97 \\
        \cmidrule(lr){2-13}
         & \multirow{3}{*}{V4} & PTFS & 48.50 & 8.19 & 24.73 & 44.37 & 26.82 & \bf 25.70 & 25.19 & 19.49 & 15.36 & 26.48 \\
         & & CPT  & \bf 49.10 & 8.34 & 25.48 & 40.60 & \bf 27.27 & 22.54 & 25.38 & 21.39 & \bf 17.44 & 26.39 \\
        \cdashline{3-13}
         & & Ours & 48.20 & \bf 9.02 & \bf 26.30 & \bf 44.56 & \bf 27.27 & 23.69 & \bf 25.56 & \bf 22.53 & 14.20 & \bf 26.81 \\
        \midrule
        \multirow{9}{*}{\parbox{0.7cm}{3.1B\\42B}} & \multirow{3}{*}{V2} & PTFS & 41.10 & 6.37 & 24.00 & 36.43 & 24.15 & 21.62 & 24.97 & 19.75 & \bf 14.22 & 23.62 \\
         & & CPT  & \bf 46.00 & 6.14 & 24.00 & \bf 40.81 & \bf 27.04 & 21.94 & 23.57 & \bf 20.89 & 13.28 & 24.85 \\
        \cdashline{3-13}
         & & Ours & 43.70 & \bf 8.57 & \bf 24.23 & 40.17 & 25.78 & \bf 24.59 & \bf 25.70 & 19.37 & \bf 14.22 & \bf 25.15\\
        \cmidrule(lr){2-13}
         & \multirow{3}{*}{V3} & PTFS & 44.30 & 8.34 & 23.83 & 40.99 & \bf 27.12 & 21.71 & 24.73 & \bf 21.65 & \bf 15.48 & 25.35 \\
         & & CPT  & 43.90 & 8.11 & \bf 25.23 & \bf 41.24 & 26.00 & 25.00 & \bf 25.44 & 20.00 & 13.40 & 25.37 \\
        \cdashline{3-13}
         & & Ours & \bf 47.90 & \bf 9.48 & 24.02 & 40.74 & 25.71 & \bf 25.73 & 25.09 & 19.62 & 14.54 & \bf 25.87 \\
        \cmidrule(lr){2-13}
         & \multirow{3}{*}{V4} & PTFS & 50.20 & \bf 11.22 & \bf 25.98 & 39.89 & 27.64 & \bf 23.12 & 25.47 & 21.65 & \bf 15.46 & 26.74 \\
         & & CPT  & \bf 50.60 & 9.78 & 25.12 & 41.03 & \bf 28.08 & 22.48 & 25.38 & 21.01 & 13.93 & 26.38 \\
        \cdashline{3-13}
         & & Ours & 49.80 & 10.77 & 25.77 & \bf 42.95 & 26.97 & 22.45 & \bf 26.25 & \bf 22.41 & 14.80 & \bf 26.91 \\
        \bottomrule
    \end{tabular}}
    \caption{The performance of downstream tasks for LLMs across four versions. In addition to the standard training scale (LLaMA-1.2B trained for 42B tokens), we further evaluate LLMs trained on more data (LLaMA-1.2B trained for 168B tokens) and LLMs with a larger size (LLaMA-3.1B trained for 42B tokens).}
    \label{tab:detailed-down-task}
\end{table*}

\begin{table}[t]
    \centering
    \begin{tabular}{c c ccc}
        \toprule
        \multirow{2}{*}{$\mathbf{\alpha}$} & \multirow{2}{*}{\bf TP} & \multicolumn{3}{c}{\bf PPL} \\
        \cmidrule(lr){3-5}
         & & \bf V2 & \bf V3 & \bf V4 \\
        \midrule
        \multirow{2}{*}{0.2} & CPT & 21.40 & 20.69 & 20.20 \\
        \cdashline{2-5}
         & Ours & \bf 20.96 & \bf 20.05 & \bf 19.59 \\
        \midrule

        \multirow{2}{*}{0.4} & CPT & 21.06 & 20.43 & 20.18 \\
        \cdashline{2-5}
         & Ours & \bf 20.81 & \bf 19.88 & \bf 19.42 \\
        \midrule

        \multirow{2}{*}{0.6} & CPT & 20.85 & 20.21 & 19.85 \\
        \cdashline{2-5}
         & Ours & \bf 20.82 & \bf 19.86 & \bf 19.40 \\
        
        \bottomrule
    \end{tabular}
    \caption{The comparison between CPT and our paradigm with equal training cost. The $\alpha$ ranges from 0.2 to 0.6.}
    \label{tab:comparison-cpt-ours}
\end{table}

\begin{table}[t]
    \centering
    \begin{tabular}{l ccc}
        \toprule
        \multirow{2}{*}{\bf TP} & \multicolumn{3}{c}{\bf PPL} \\
        \cmidrule(lr){2-4}
         & \bf 10.5B & \bf 21.0B & \bf 31.5B \\
        \midrule
        PTFS & 20.84 & 19.28 & 18.36 \\
        CPT  & 21.11 & 19.50 & 18.47  \\
        \cdashline{1-4}
        Ours & \bf 20.13 & \bf 18.81 & \bf 18.09 \\
        \bottomrule
    \end{tabular}
    \caption{The comparison of different paradigms for training two versions of LLaMA-1.2B. The data increment of the second version varies from 10.5B to 31.5B tokens.}
    \label{tab:diff-paradigm-1.2B_diff-data-increment}
\end{table}

\section{Comparison between CPT and Ours}
\label{sec:cpt_ours}

Existing experimental results show that while our paradigm outperforms CPT in terms of performance, it has higher training cost. 
To provide a more direct comparison between CPT and our paradigm, we conduct an experiment where the training cost (measured by training steps) are kept consistent.
Concretely, we sample a dataset of approximately 5.25B tokens (5K steps) and use it to train four versions of LLMs. 
As mentioned in section~\ref{subsec:complexity-analysis}, we analyze the time complexity of CPT and our paradigm, obtaining that the time complexity of ours is about $1+\alpha$ times that of CPT.

We compare the results for different $\alpha$ (proportion of fast-decaying steps) set as $0.2$, $0.4$ and $0.6$, respectively. 
To ensure that the total training cost used for these two paradigms are consistent, our paradigm always includes an additional 10K steps for each version update, while CPT uses additional 12K ($\alpha=0.2$), 14K ($\alpha=0.4$) and 16K ($\alpha=0.6$) steps for each version update, respectively. 
The experimental results in Table~\ref{tab:comparison-cpt-ours} demonstrate that our paradigm remains effective even when the total training cost are kept consistent with CPT.

\section{Version Updates with Inconsistent Data Increments}
\label{sec:inconsistent_data}

Existing experiments are based on the assumption of consistent data increments during version updates of LLMs.
The effectiveness of our paradigm has not yet been validated in the scenario with varying data increments.
Hence, we conduct a comparative experiment involving PTFS, CPT, and ours, training two versions of LLMs.
For all paradigms, the LLMs of the first version are trained with 10.5B tokens (10K steps), while the LLMs of second version are trained with 10.5B, 21B, and 31.5B tokens, respectively.
As the experimental results shown in Table~\ref{tab:diff-paradigm-1.2B_diff-data-increment}, our paradigm maintains a better pre-training performance than PTFS and CPT in the scenario with inconsistent data increments. 
This further demonstrates the generalization of our paradigm.

\end{document}